\documentclass[conference,a4paper]{APSIPA2025}
\usepackage{graphicx}
\usepackage{amsmath}
\usepackage{amssymb}
\usepackage{amsfonts}
\usepackage{amsxtra}
\usepackage{mathrsfs}
\usepackage{latexsym}
\usepackage{arydshln}
\usepackage{url}
\usepackage{flushend}
\usepackage{algorithm}
\usepackage[noend]{algpseudocode}
\usepackage{xcolor}
\usepackage{listings}
\usepackage{multirow}
\usepackage{booktabs}

\usepackage{mathtools}
\usepackage{cleveref}

\makeatletter
\let\MYcaption\@makecaption
\makeatother
\usepackage{subcaption}
\makeatletter
\let\@makecaption\MYcaption
\makeatother
\renewcommand{\vec}[1]{\boldsymbol{#1}}
\newcommand{\mat}[1]{\boldsymbol{\mathrm{#1}}}
\newcommand{\set}[1]{\mathbb{#1}}

\newcounter{num}

\newcommand{\ip}[2]{#1^\top \: #2}

\DeclareMathOperator*{\argmax}{arg~max}

\crefname{equation}{Eq.}{Eqs.}% {環境名}{単数形}{複数形} \crefで引くときの表示
\crefname{figure}{Fig.}{Figs.}% {環境名}{単数形}{複数形} \crefで引くときの表示
\crefname{table}{Table}{Tables}% {環境名}{単数形}{複数形} \crefで引くときの表示
\crefname{algorithm}{Algorithm}{Algorithm}
\usepackage{geometry}
\usepackage{fancyhdr}

\fancypagestyle{firststyle}{
  \fancyhf{}
  \fancyhead[C]{2025 Asia Pacific Signal and Information Processing Association Annual Summit and Conference (APSIPA ASC)}
}

\begin{document}

\title{Scale and Rotation Estimation of Similarity-Transformed Images via Cross-Correlation Maximization Based on Auxiliary Function Method}

\author{
\authorblockN{
Shinji Yamashita\authorrefmark{1},
Yuma Kinoshita\authorrefmark{1}\authorrefmark{2}, and
Hitoshi Kiya\authorrefmark{2}
}

\authorblockA{
\authorrefmark{1}
Tokai University, Japan\\
}
\authorblockA{
\authorrefmark{2}
Tokyo Metropolitan University, Japan\\
}
}
\maketitle
\thispagestyle{firststyle}
\pagestyle{fancy}
\cfoot{ }

\begin{abstract}
This paper introduces a highly efficient algorithm capable of jointly estimating scale and rotation between two images with sub-pixel precision.
Image alignment serves as a critical process for spatially registering images captured from different viewpoints, and finds extensive use in domains such as medical imaging and computer vision.
Traditional phase-correlation techniques are effective in determining translational shifts;
however, they are inadequate when addressing scale and rotation changes,
which often arise due to camera zooming or rotational movements.
In this paper, we propose a novel algorithm that integrates scale and rotation estimation based on the Fourier transform in log-polar coordinates with a cross-correlation maximization strategy,
leveraging the auxiliary function method.
By incorporating sub-pixel-level cross-correlation
our method enables precise estimation of both scale and rotation.
Experimental results demonstrate that the proposed method achieves lower mean estimation errors for scale and rotation than conventional Fourier transform-based techniques that rely on discrete cross-correlation.

\end{abstract}

\section{Introduction}
  Image alignment is a technique that spatially registers images
  by correcting geometric transformations between them.
  This technique is essential in a wide range of applications,
  including medical image processing~\cite{medical}, panorama stitching~\cite{panorama},
  3D reconstruction~\cite{3D}, and high dynamic range (HDR) image generation~\cite{hdr1, hasinoff2016burst}.
  In recent years, there has been an increasing demand for fast and accurate alignment even on resource-constrained devices such as smartphones, highlighting the importance of low-cost algorithms.

  There are two principal approaches to image alignment: homography-based methods and intensity-based methods. Homography-based approaches estimate a homography matrix between two images by leveraging feature descriptors such as SIFT~\cite{lowe2004distinctive}, SURF~\cite{bay2006surf}, and A-KAZE~\cite{alcantarilla2012kaze}, often in conjunction with the RANSAC algorithm~\cite{fischler1981random}. Recently, deep learning-based techniques for homography estimation have also been investigated~\cite{cao2022iterative,hong2022unsupervised}. Intensity-based methods generally focus on estimating optical flow~\cite{lucas1981iterative,dosovitskiy2015flownet,ilg2017flownet,jung2023anyflow}. However, both homography and optical flow estimation methods typically incur considerable computational costs.

  To address these limitations,
  two-dimensional (2D) cross-correlation-based methods
  ~\cite{knapp1976generalized,takita2003highaccuracy} are frequently employed for real-time image alignment~\cite{hasinoff2016burst}.
  These methods aim to maximize the 2D cross-correlation function between two images to estimate the translational displacement.
  The cross-correlation for discrete pixel shifts can be computed efficiently using the fast Fourier transform (FFT), and its maximum can be readily identified.
  Furthermore, Kinoshita et al.\ proposed an efficient algorithm for maximizing the 2D cross-correlation function with subpixel accuracy, based on the auxiliary function method, also known as majorization-minimization (MM)~\cite{aux}.

  Nevertheless, methods based on maximizing the 2D cross-correlation function typically assume that the displacement between images is purely translational and estimate only the translation parameters.
  In practice, however, images often exhibit differences in scale and rotation due to camera zoom or rotation, resulting in misalignments beyond simple translation.

  In response to this issue,
  we aim to estimate scale and rotation changes with high accuracy from a pair of images including scale, rotation, and translation differences.
  To achieve this goal, we combine a Fourier transform-based scale and rotation estimation~\cite{reddy1996fftbased} with the cross-correlation maximization algorithm proposed by Kinoshita et al~\cite{aux}.
  In this method, scale and rotation are estimated by maximizing the 2D cross-correlation
  between the amplitude spectra represented in log-polar coordinate system.
  For maximizing the 2D cross-correlation,
  the use of the algorithm developed by Kinoshita et al.
  enables us to estimate scale and rotation with high accuracy.
  
  We evaluate the estimation accuracy of the proposed method and
  the baseline estimation with the standard discrete cross-correlation maximization.
  Experimental results demonstrate that the proposed method outperforms
  the baseline method in terms of absolute errors between the ground truth and the estimated values.

\section{Preliminaries}
In this section, we briefly formalize the image alignment problem and present the mathematical framework that underpins the proposed methods.

\subsection{Problem Statement}
  This paper focuses on the task of aligning two-dimensional discrete signals, particularly images, represented by $x$ and $y$. The pixel value at position $\vec{p} = (p_1, p_2)^\top$ is denoted by $x[\vec{p}]$ and $y[\vec{p}]$, where the superscript $\top$ indicates the transpose of a vector or matrix.
  If the spatial misalignment between the images $x$ and $y$ can be described by a combination of translation $\Delta \vec{p} \in \set{R}^2$, rotation $\mat{R}(\theta)$, and scaling $s \in \set{R}^+$, then the relationship can be formulated as
  \begin{equation}
    y[\vec{p}] = x[s \mat{R}(\theta) \vec{p} + \Delta \vec{p}],
    \label{eq:misalignment_model}
  \end{equation}
  where
  \begin{equation}
    \mat{R}(\theta) =
    \begin{pmatrix}
      \cos\theta & \sin\theta \\
      -\sin\theta & \cos\theta
    \end{pmatrix}.
  \end{equation}
  The aim of this paper is to estimate the parameters $\theta$, and $s$
  from the given pair $(x, y)$
  because once scale and rotation are aligned the parameter $\Delta \vec{p}$
  can be easily estimated by cross-correlation methods.

\subsection{Maximization of 2D cross-correlation}\label{sec:correlation}
  When the displacement between the images is due solely to translation (i.e., $s=1$ and $\theta=0$), the parameter $\Delta \vec{p}$ can be determined by maximizing the generalized two-dimensional cross-correlation function.
  Let $\hat{x}$ and $\hat{y}$ denote the two-dimensional discrete Fourier transforms of $x$ and $y$, respectively, both of size $N \times M$.
  We assume $\hat{x}$ and $\hat{y}$ are strictly band-limited,
  meaning $\hat{x}(\vec{\omega}) = \hat{y}(\vec{\omega}) = 0$
  for angular frequency $\vec{\omega} = (\omega_1, \omega_2)^\top$
  where at least one component equals $\pi$.
  Utilizing the 2D cross-spectrum is defined as 
  $\hat{\Phi}_2^{(xy)}(\vec{\omega}_{kl}) = \hat{x}^{*}(\vec{\omega}_{kl}) \hat{y}(\vec{\omega}_{kl})$,
  the generalized two-dimensional cross-correlation function between $x$ and $y$ is given by
  \begin{equation}
    \check{\Phi}_2^{(xy)}[\vec{p}]
    = \frac{1}{NM}
      \sum_{k \in K} \sum_{l \in L}
        w_{kl} \hat{\Phi}_2^{(xy)}(\vec{\omega}_{kl})
        \exp \left(j \ip{\vec{\omega}_{kl}}{\vec{p}} \right),
    \label{eq:gcc2}
  \end{equation}
  where $j$ donates the imaginary unit,
  $w_{kl} \in \set{R}^+$ denotes an arbitrary positive weighting coefficient,,
  and $\vec{\omega}_{kl} = (\omega^{(k)}, \omega^{(l)})^\top = (\frac{2 \pi k}{N}, \frac{2 \pi l}{M})^\top$.
  The sets $K$ and $L$ are defined as
  $K = \{-N/2+1, -N/2+2, \cdots, N/2\}$ and $L = \{-M/2+1, -M/2+2, \cdots, M/2\}$,
  respectively.
  When $w_{kl} = 1$,
  the function reduces to the standard cross-correlation.
  In the case where $w_{kl}=|\hat{\Phi}_2^{(xy)}(\vec{\omega}_{kl})|^{-1}$,
  the method becomes equivalent to phase-only correlation (POC)~\cite{kuglin1975phase},
  which is also referred to as generalized cross-correlation with phase transform (GCC-PHAT)~\cite{knapp1976generalized} in the context of acoustic signal processing.

  Kinoshita et al.~\cite{aux} proposed an algorithm that regards $\vec{p}$ in \cref{eq:gcc2} as a real-valued vector,
  considers the continuous function $\Phi_2^{(xy)}$,
  and maximizes
  \begin{equation}
    \tilde{\Delta \vec{p}} = \argmax_{\vec{p} \in \set{R}^2} \Phi_2^{(xy)}(\vec{p})
    \label{eq:objective}
  \end{equation}
  using the auxiliary function method, thereby achieving fast convergence to a local optimum and subpixel accuracy in displacement estimation.
  Instead of directly maximizing the objective function in \cref{eq:objective}, the algorithm iteratively maximizes an auxiliary function $Q(\vec{p}, \theta)$ that serves as its lower bound.
  \begin{equation}
    \vec{\theta}^{(i)} = f(\vec{p}^{(i)}), \quad
    \vec{p}^{(i+1)} = \argmax_{\vec{p} \in \set{R}^2} Q(\vec{p}, \vec{\theta}^{(i)}).
  \end{equation}
  Here, $i$ denotes the iteration number.

\subsection{Fourier Transform-Based Scale and Rotation Estimation}\label{sec:estimation}
The spectra $\hat{x}$ and $\hat{y}$ of the signals
$x(\vec{p})$ and $y(\vec{p}) = x(s\mat{R}(\theta)\vec{p}+\Delta \vec{p})$
satisfy the following relationship:
\begin{equation}
  \hat{y}(\vec{\omega})
  = \exp\!\bigl(j 2\pi \vec{\omega}^{\top}\mat{R}(-\theta)\Delta\vec{p}\bigr)\,
    \frac{1}{s^2}\,
    \hat{x}\!\left(\frac{1}{s} \mat{R}(\theta) \vec{\omega}\right).
    \label{eq:fourier_similarity}
\end{equation}
Consequently, the amplitude spectrum is invariant to translation and, after appropriate scaling and rotation, coincides with the amplitude spectrum of the other signal.
When the amplitude spectrum is represented in log-polar coordinates $(\rho, \phi)$, where $\rho = \log \| \vec{\omega} \|$ and $\phi = \tan^{-1} ( \omega_2 / \omega_1 )$, scaling corresponds to a translation along the $\rho$ axis, whereas rotation corresponds to a translation along the $\phi$ axis.
By exploiting this property, the scale factor $s$ and rotation angle $\theta$ can be estimated by maximizing the cross-correlation between the amplitude spectra in the log-polar domain~\cite{reddy1996fftbased}.
The derivation of \cref{eq:fourier_similarity} is presented below.

When $y(\vec{p})$ is an affine transformation of $x(\vec{p})$,
that is,
$y(\vec{p}) = x(\mat{A}\vec{p} + \vec{b})$ with $\mat{A} \in \mathbb{R}^{2\times2}$ and $\vec{b} \in \mathbb{R}^2$,
the Fourier transform is given by
\begin{align}
  \hat{y}(\vec{\omega})
  &= \iint_{\mathbb{R}^2} y(\vec{p}) \exp(-j 2\pi \vec{\omega}^\top \vec{p})\, d\vec{p} \\
  &= \iint_{\mathbb{R}^2} x(\mat{A} \vec{p} + \vec{b}) \exp(-j 2\pi \vec{\omega}^\top \vec{p})\, d\vec{p}
\end{align}
By applying the change of variables $\vec{q} = \mat{A} \vec{p} + \vec{b}$ yields $\vec{p} = \mat{A}^{-1} (\vec{q} - \vec{b})$, and therefore,
\begin{equation}
  d\vec{p} = \left| \frac{\partial \vec{p}}{\partial \vec{q}} \right|\, d\vec{q} = \frac{1}{|\mat{A}|} d\vec{q}.
\end{equation}
Thus,
\begin{align}
  \hat{y}(\vec{\omega}) &= \frac{1}{|\mat{A}|} \iint_{\mathbb{R}^2} x(\vec{q}) \exp\left(-j 2\pi \vec{\omega}^\top \mat{A}^{-1} (\vec{q} - \vec{b})\right)\, d\vec{q} \\
  &= \exp\left(j 2\pi \vec{\omega}^\top \mat{A}^{-1} \vec{b}\right) \frac{1}{|\mat{A}|} \hat{x}\left((\mat{A}^{-1})^\top \vec{\omega}\right), \label{eq:fourier_affine}
\end{align}
where
\begin{equation}
\hat{x}\left((\mat{A}^{-1})^\top \vec{\omega}\right) = \iint_{\mathbb{R}^2} x(\vec{q}) \exp\left(-j 2\pi \vec{\omega}^\top \mat{A}^{-1} \vec{q}\right) d\vec{q}.
\end{equation}
By substituting $\mat{A} = s \mat{R}(\theta)$ and $\vec{b} = \Delta \vec{p}$ into \cref{eq:fourier_affine}, we recover the expression in \cref{eq:fourier_similarity}.

In this paper, we combine the Fourier transform-based scale and rotation estimation with the cross-correlation maximization algorithm proposed by Kinoshita et al.,
enabling highly accurate estimation of the rotation angle $\theta$ and the scale factor $s$.

\section{Proposed Method}
In this section, we present our algorithm for estimating the displacement in scale and rotation between two images, $x$ and $y$.

\subsection{Overview}
\begin{figure}[t]
    \centering
    \includegraphics[width=1\linewidth]{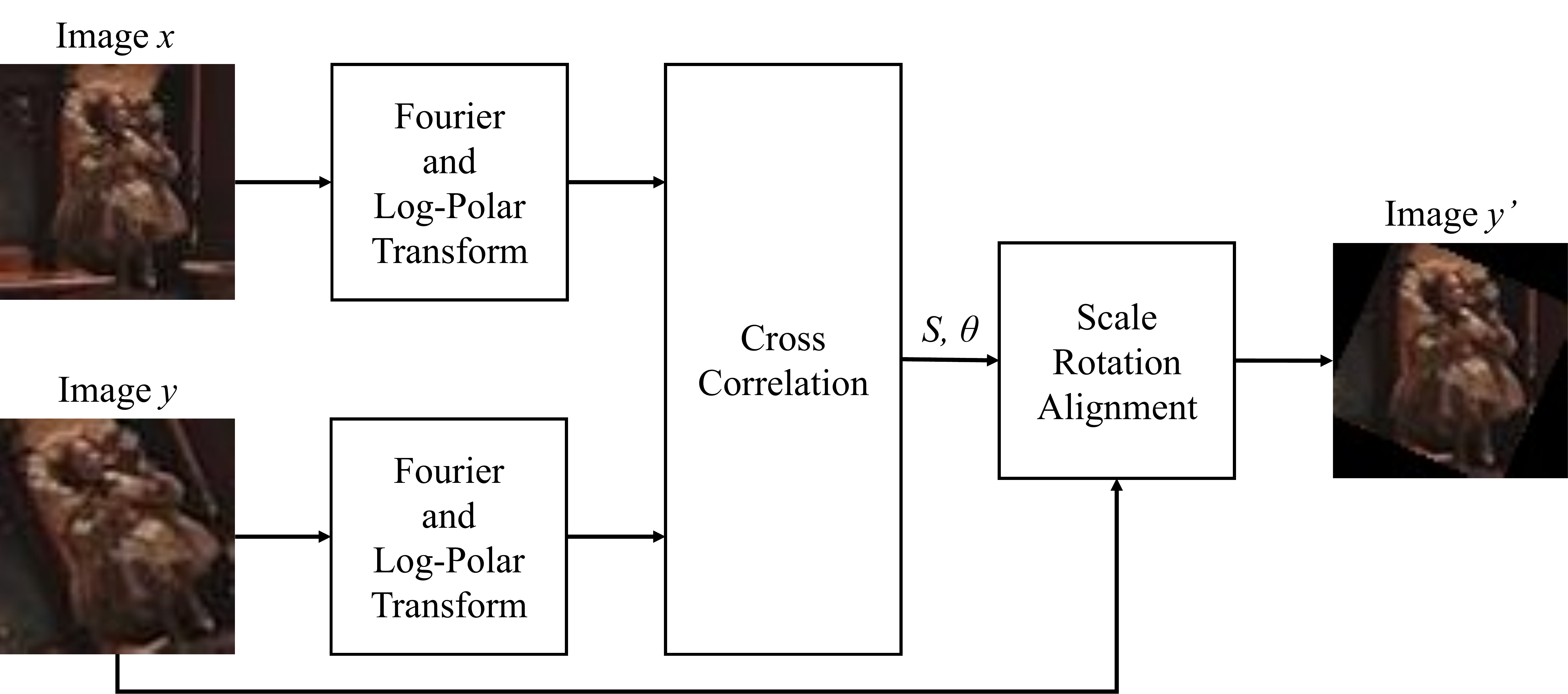}
    \caption{Procedure of proposed method}
    \label{flowchart}
\end{figure}

The procedure for the proposed method is illustrated in \cref{flowchart}.
To estimate the displacement in scale and rotation, we maximize the cross-correlation between the amplitude spectra of the images represented in log-polar coordinates, since scaling and rotation correspond to translation in this transformed domain, as described in \cref{sec:estimation}.
Here, we adopt the continuous cross-correlation and iteratively optimize its auxiliary function, as discussed in \cref{sec:correlation}.

The proposed procedure consists of the following steps:
\begin{enumerate}
  \item Compute the amplitude spectra $|\hat{x}|$ and $|\hat{y}|$ of the images $x$ and $y$, respectively.
  \item Transform the amplitude spectra into log-polar coordinates.
  \item Estimate the scale and rotation parameters by maximizing the continuous cross-correlation between the spectra in the log-polar domain.
\end{enumerate}
Each step is explained in detail below.

\subsection{Calculation of Amplitude Spectrum}
To obtain the amplitude spectrum of image $x$,
we compute the discrete Fourier transform using a window function $\varpi$,
\begin{equation}
  \hat{x}[\vec{\omega}]
  = \sum_{p_1=0}^{N-1} \sum_{p_2=0}^{M-1}
      \varpi[\vec{p}] x[\vec{p}] \exp \left(-j \ip{\vec{\omega}}{\vec{p}} \right),
  \label{eq:dft}
\end{equation}
where we use a 2D Gaussian window with parameters $\sigma_1 = N/5$ and $\sigma_2 = M/5$ as $\varpi$.
The amplitude $|\hat{x}|$ is then computed.
The amplitude spectrum of image $y$ is calculated in the same manner.

\subsection{Log-Polar Transform}
The amplitude spectra $|\hat{x}|$ and $|\hat{y}|$ in Cartesian coordinates
are transformed into log-polar coordinates.
This transformation is referred to as the log-polar transform.
The log-polar transformation is defined for $\vec{r} = (\rho, \phi)^\top$ as follows:
\begin{equation}
  |\hat{X}[\vec{r}]| = |\hat{x}[(e^\rho \cos \phi, e^\rho \sin \phi)^\top]|,
  \label{eq:log_polar}
\end{equation}
where $\rho = \log \| \vec{\omega} \|$ and $\phi = \tan^{-1}(\omega_2/\omega_1)$.
We evaluate \cref{eq:log_polar} at discrete points
$\rho \in \{\frac{\log(\min(N, M))}{N}i : i = 1, \ldots, N\}$ and $\phi \in \{\frac{2\pi}{M} i : i = 0, \ldots, M-1\}$.
Since $|\hat{x}|$ is a discrete signal,
bi-cubic interpolation is applied to $|\hat{x}|$ for resampling it.
The amplitude spectrum $|\hat{Y}|$ of image $|\hat{y}|$ is computed in the same manner.

\subsection{Estimation of Scale and Rotation}
The cross-correlation function between $|\hat{X}|$ and $|\hat{Y}|$ is maximized
using the algorithm of Kinoshita et al., as described in \cref{sec:correlation}.
When calculating the cross-spectrum $\hat{\Phi}^{(XY)}$ of $|\hat{X}|$ and $|\hat{Y}|$,
both $|\hat{X}|$ and $|\hat{Y}|$ are zero-mean normalized and processed with the Gaussian window $\varpi$.
As a result, the estimated translational displacement
$\tilde{\Delta \vec{r}} = (\tilde{\rho}, \tilde{\phi})^\top$
in the log-polar domain yields the estimates for scale and rotation, 
$\tilde{s}$ and $\tilde{\theta}$, as follows:
\begin{equation}
    \tilde{s} =  \exp \left(\frac{\log(\min(N, M))}{N} \tilde{\rho}\right), \quad \tilde{\theta} = \frac{2\pi}{M}\tilde{\phi} 
\end{equation}

\section{Experiments}
To evaluate the performance of the proposed method in terms of estimation accuracy,
we conducted a simulation experiment.

\subsection{Experimental Setup}
\begin{figure}[t]
     \centering
     \includegraphics[width=1\linewidth]{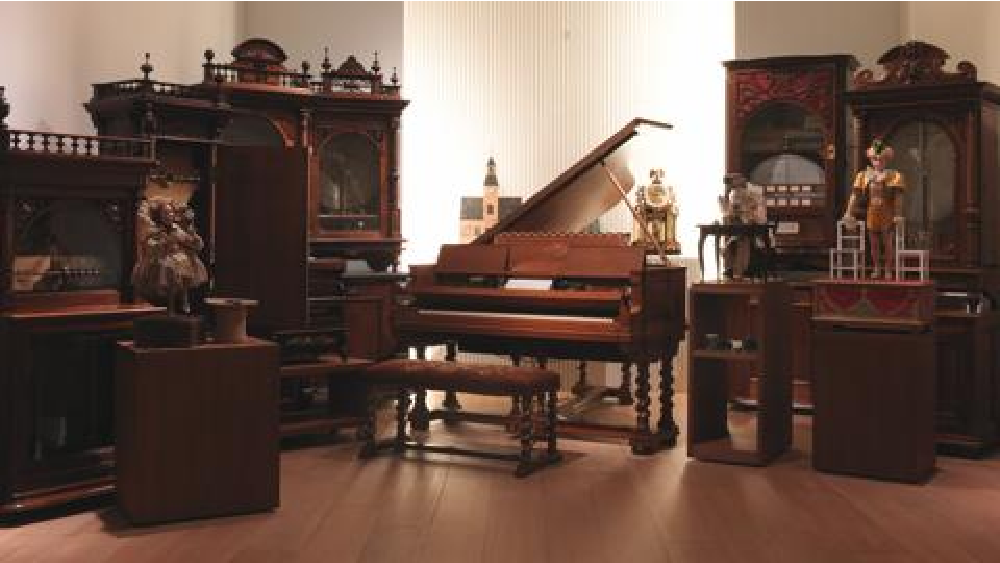}
     \caption{Original high-resolution image “MusicBox” used in the experiment}
     \label{MusicBox}
 \end{figure}

Five pairs of images were prepared by simulating similarity transformations.
In this experiment, random similarity transformations with parameters $s$, $\theta$, and $\Delta \vec{p}$
were applied to a high-resolution image, "MusicBox" (shown in Fig.~\ref{MusicBox}),
which was obtained from the “Ultra-high Definition/Wide-Color-Gamut Standard Images” dataset\footnote{\url{https://www.ite.or.jp/content/test-materials/uhdtv/}}.
Image pairs were subsequently generated by randomly cropping $64 \times 64$ pixel regions
from the both original and the transformed images at corresponding locations.

The parameters, namely the scaling factor $s$, rotation angle $\theta$, and translation vector $\Delta \vec{p}$, were randomly sampled from uniform distributions over the intervals listed in \cref{tab:parameter_range}.
\begin{table}[t]
  \centering
  \caption{Ranges of transformation parameters used in the experiment.\label{tab:parameter_range}}
  \begin{tabular}{l|c}
    \hline
    Parameter & Range \\
    \hline
    Rotation angle $\theta \ ({}^\circ)$ & -30 -- 30 \\
    Scaling factor $s$ & 0.8  -- 1.2 \\
    Horizontal translation $p_1$ (px.) & -5 -- 5\\ 
    Vertical translation $p_2$ (px.)& -5 -- 5\\
    \hline
  \end{tabular}
\end{table}
The absolute errors between the ground-truth and the estimated values
were computed to evaluate the estimation accuracy.

The proposed method was compared against the conventional Fourier transform-based
scale and rotation estimation approach, which maximizes the standard discrete cross-correlation (Baseline).

\subsection{Experimental Results}
The experimental results are summarized in Table~\ref{result}.
For each method, the table reports the estimation errors in scale and rotation angle for five image pairs, along with their corresponding averages and variances.
\begin{table}[t]
\centering
\caption{Absolute estimation errors for scale and rotation angle: comparison between the baseline and proposed methods}
\label{result}
\begin{tabular}{lcccc} \toprule
               & \multicolumn{2}{c}{Baseline}                      & \multicolumn{2}{c}{Ours}                          \\
               & Scale & Angle & Scale & Angle \\ \midrule
Image 1        & 0.038 & \textbf{0.676}          & \textbf{0.031} & 0.924 \\
Image 2        & 0.002 & 1.987          & \textbf{0.001} & \textbf{0.702} \\
Image 3        & 0.193 & 2.085          & \textbf{0.192} & \textbf{0.472} \\
Image 4        & 0.074 & \textbf{1.430}          & \textbf{0.040} & 3.095 \\
Image 5        & 0.072 & 2.281          & \textbf{0.012} & \textbf{0.783} \\ \midrule
Average        & 0.076 & 1.692           & \textbf{0.055} & \textbf{1.195} \\
Variance       & 0.003 & 0.281          & 0.004 & 0.769 \\ \bottomrule
\end{tabular}
\end{table}
As shown in Table~\ref{result},
the proposed method consistently achieves lower average errors in both scale and rotation angle estimation than the baseline method,
indicating improved estimation accuracy.

\begin{figure}[!t]
    \centering
    \begin{subfigure}[t]{0.45\hsize}
      \centering
      \includegraphics[width=\linewidth]{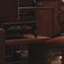}
      \caption{Input image $x$}
      \label{Image 2: reference image}
    \end{subfigure}
    \begin{subfigure}[t]{0.45\hsize}
      \centering
      \includegraphics[width=\linewidth]{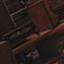}
      \caption{Input image $y$}
      \label{Image 2: before correct}
    \end{subfigure}\\ \vspace{1em}
    \begin{subfigure}[t]{0.45\hsize}
      \centering
      \includegraphics[width=\linewidth]{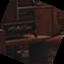}
      \caption{Baseline}
      \label{Image 2: dcc}
    \end{subfigure}
    \begin{subfigure}[t]{0.45\hsize}
      \centering
      \includegraphics[width=\linewidth]{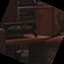}
      \caption{Ours}
      \label{Image 4}
    \end{subfigure}
    \caption{Alignment results for Image 2\label{fig:image2}}
\end{figure}

\begin{figure}[!t]
    \centering
    \begin{subfigure}[t]{0.45\hsize}
      \centering
      \includegraphics[width=\linewidth]{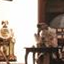}
      \caption{Input image $x$}
      \label{Image 2: reference image}
    \end{subfigure}
    \begin{subfigure}[t]{0.45\hsize}
      \centering
      \includegraphics[width=\linewidth]{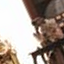}
      \caption{Input image $y$}
      \label{Image 2: before correct}
    \end{subfigure}\\ \vspace{1em}
    \begin{subfigure}[t]{0.45\hsize}
      \centering
      \includegraphics[width=\linewidth]{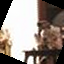}
      \caption{Baseline}
      \label{Image 2: dcc}
    \end{subfigure}
    \begin{subfigure}[t]{0.45\hsize}
      \centering
      \includegraphics[width=\linewidth]{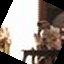}
      \caption{Ours}
      \label{Image 4}
    \end{subfigure}
    \caption{Alignment results for Image 3\label{fig:image3}}
\end{figure}

For Images 2 and 3, the input images $x$ and $y$, as well as the alignment results
based on scale and rotation estimation by both the baseline and proposed methods,
are illustrated in \cref{fig:image2,fig:image3}, respectively.
As seen in \cref{fig:image2}, both the baseline and the proposed methods achieve accurate scale alignment.
Furthermore, the proposed method demonstrated superior rotation estimation
compared to the baseline, as reflected in Table~\ref{result}.
As seen in \cref{fig:image3}, although the proposed method outperforms
the baseline in terms of rotation estimation,
both methods yield suboptimal estimations of the scale factor, as also indicated in Table~\ref{result}.

These results indicate that while the proposed method generally provides higher estimation accuracy
than the baseline, there remain certain images for which the estimation is not sufficiently reliable.
Achieving robust and highly accurate estimation for arbitrary input images remains an important area
for future work.

\section{Conclusions}
In this paper, we have proposed a novel approach that combines
Fourier transform-based scale and rotation estimation
with the cross-correlation maximization algorithm developed by Kinoshita et al.
The integration of Kinoshita's algorithm enables highly accurate estimation of
both scale and rotation parameters.
Experimental results demonstrated that the proposed method generally achieves superior estimation
accuracy compared to the baseline method;
however, suboptimal estimations were observed for certain image pairs.

For future work, we aim to further enhance estimation accuracy across a wider variety of input images, and to extend the alignment framework to simultaneously estimate translation parameters
in addition to scale and rotation.

%\printbibliography
%\bibliographystyle{IEEEtran}
%\bibliography{mybib}
% Generated by IEEEtran.bst, version: 1.14 (2015/08/26)

\end{document}